# AUTOMATIC PATTERN CLASSIFICATION BY UNSUPERVISED LEARNING USING DIMENSIONALITY REDUCTION OF DATA WITH MIRRORING NEURAL NETWORKS

Name(s) Dasika Ratna Deepthi [(1)], G.R.Aditya Krishna [(2)] and K. Eswaran [(3)]
Address: (1) Ph. D. Student, College of Engineering, Osmania University, Hyderabad-500007, A.P.,
(2) Altech Imaging & Computing, Sri Manu Plaza, A.S. Rao Nagar, Hyderabad, 500062, A.P.,
(3) Sreenidhi Institute of Science and Technology, Yamnampet, Ghatkesar, Hyderabad - 501 301. A.P.,
Country India
Email addresses: radeep07@gmail.com, adityagr@gmail.com, kumar.e@gmail.com

**ABSTRACT**

*This paper proposes an unsupervised learning technique by using Multi-layer Mirroring Neural Network and Forgy's clustering algorithm. Multi-layer Mirroring Neural Network is a neural network that can be trained with generalized data inputs (different categories of image patterns) to perform non-linear dimensionality reduction and the resultant low-dimensional code is used for unsupervised pattern classification using Forgy's algorithm. By adapting the non-linear activation function (modified sigmoidal function) and initializing the weights and bias terms to small random values, mirroring of the input pattern is initiated. In training, the weights and bias terms are changed in such a way that the input presented is reproduced at the output by back propagating the error. The mirroring neural network is capable of reducing the input vector to a great degree (~ $1/30^{th}$ the original size) and also able to reconstruct the input pattern at the output layer from this reduced code units. The feature set (output of central hidden layer) extracted from this network is fed to Forgy's algorithm, which classify input data patterns into distinguishable classes. In the implementation of Forgy's algorithm, initial seed points are selected in such a way that they are distant enough to be perfectly grouped into different categories. Thus a new method of unsupervised learning is formulated and demonstrated in this paper. This method gave impressive results when applied to classification of different image patterns.*

From now on Multi-layer Mirroring Neural Network will be abbreviated as MMNN.

## 1. Introduction

Feature extraction of patterns in data and the task of dimension reduction, that is, choosing as few parameters as possible for the characterization of data, are important steps in computer aided pattern recognition. This may be supervised or unsupervised depending on whether or not the pattern label information is provided at the time of training and classification. This paper deals with extracting feature set from the input using MMNN and unsupervised classification using Forgy's clustering algorithm. There are many techniques for features extraction of the object and machine learning [1], [2] which can be applied to recognition. An elaborate discussion on several techniques used to perform nonlinear dimensionality reduction can be found in [3]. [4] and [5] deal with the neural network approach to dimensionality reduction. For dimensionality reduction there are other techniques like PCA, ICA**.** In [6], [7] and [8] dimensionality reduction is used for visualization of patterns in data using PCA and multi-dimensional scaling. According to the discussion given in [9] neural network approach performs better than PCA.

This paper deals with
- Nonlinear dimensionality reduction (Feature extraction)
- Reconstruction of the input patterns
- Classification of data in unsupervised mode

For dimension reduction and reconstruction MMNN [10] is used and to classify data in an unsupervised mode Forgy's algorithm is used.

MMNN is a generalized network which can take different input patterns and produce their mirrors (same as input) as output. At the same time, it also reduces the dimension of input data. This is a very simple generalized network where there is no need to change the network architecture as training proceeds (as done in other neural network approach [11]). This approach has good ability to learn and reconstruct image patterns having extremely different features fed to the network. With this approach we report a 95% success rate in the classification of different patterns.

## 2. Pattern Classification by Unsupervised Learning

The algorithm presented here is an attempt of classifying different data patterns using mirroring neural network and Forgy's unsupervised classification algorithm.

We first use the mirroring neural network to reduce the dimensions of input data pattern. The reduced dimension feature vectors thus obtained are then classified by using Forgy's unsupervised classification algorithm. Thus we obtain an algorithm which not only performs dimension reduction of the features but also clusters similar data patterns into one group.

In this work, we present an MMNN having compatible number of nodes at the first hidden layer to participate in the learning process by accepting input patterns from the input layer. This network is designed to have least possible number of nodes at central hidden layer to reduce the dimension of input pattern the output layer has same number of nodes as input layer and is used to reconstruct (mirror) the input data pattern. Using least dimensional central hidden layer outputs as input feature vector for Forgy's algorithm, data patterns are grouped into different clusters. In contrast to the typical training approach where the neural network accepts categorized input pattern, this algorithm learns different patterns and categorize them into groups by itself.

A pictorial representation of MMNN architecture is given in Fig. 1. MMNN architecture resembles the architecture given in [12]. The difference between these two is that in [12], the architecture is designed with specialized mirroring networks for each category of the input pattern, where as in MMNN, the network is a generalized network that can accept any type of input pattern. Using Forgy's unsupervised clustering algorithm input pattern is categorized into one of the different groups.

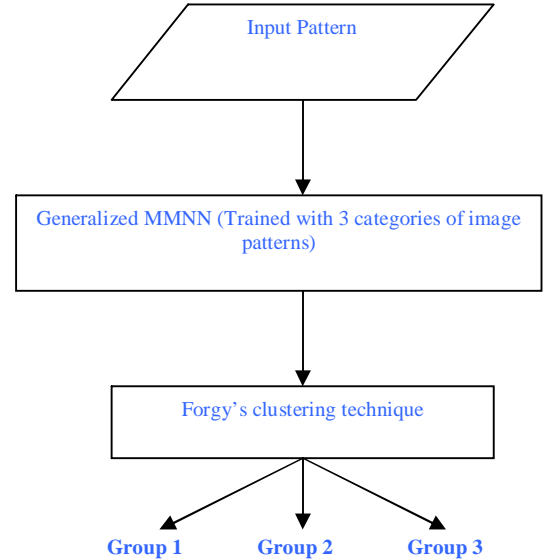

Fig. 1 MMNN that accepts generalized input patterns and classifies them in to one of the three categories with the help of Forgy's algorithm

### 2.1 Inputs to the Multi-layer Mirroring Neural Network

Inputs to MMNN are three different categories of 8-Bit images of size 26X26. These include faces, flowers and furniture. MMNN uses hyperbolic tangent function (sigmoidal function) that accepts input from -1 to +1. To fit in to this range input intensities are linearly rescaled [13] from [0,255] to [-1, +1].

$I_{input} = (I_{input} - 128) / 128$

Where $I_{input}$ is input pixel intensity.

To accelerate the learning process a truncation function (as discussed in [14]) is implemented at the input layer.

Truncation function is defined as:

$f_{trc}(I_{input})$   $= -0.9$   if $I_{input} \leq (-0.9)$
      $= +0.9$   if $I_{input} \geq (+0.9)$
      $= I_{input}$   otherwise

### 2.2 MMNN Architecture

The mirroring neural network is trained with different categories of image patterns rescaled according to the discussion given in 2.1. Fixing the number of layers and the number of nodes in each layer is done after considerable experimentation. The first hidden layer of size 30 followed by the second hidden layer of size 20 and an output layer of size equal to the input was found to be the most suitable architecture for our network. So, we have chosen 676-30-20-676 multilayer mirroring network architecture which is in contrast to the symmetrical dimensions of the encoder and decoder layers of the "autoencoder" described in [9] and "auto-associative" neural networks discussed in [3]. With this, it is apparent that there exists a nonlinear relationship between the dimensions of adjacent layers (excluding the input layer) to accurately reconstruct the input at the output of the network. We modified the sigmoidal activation function and used hyperbolic tangent function, instead of logistic function implemented in [9], as it was found to be appropriate for our input data as described in [15]. The functional output at each node of the hidden layers and output layer with activation function of the form $Sgm(s) = \tanh(s/2)$ passed through modified hyperbolic tangent function as discussed below.

The output y, which has to be passed through the filter that produces modified sigmoidal function having the upper and lower bounds at +0.9 and -0.9 respectively, is given by:

$$f_{mod}(y) = \begin{cases} -0.9 & \text{if } y \leq (-0.9) \\ +0.9 & \text{if } y \geq (+0.9) \\ y & \text{otherwise} \end{cases}$$

Where y is the sigmoidal hyperbolic tangent function defined as:

$$y = Sgm(s) = \tanh(s/2) = (1 - e^{-s}) / (1 + e^{-s})$$

Modified Hyperbolic Tangent Function forces the output of the multilayer mirroring neural network in the range of -0.9 and +0.9 instead of -1.0 and +1.0. This prevents the multilayer neural network from out-of-range values and helps in faster convergence. MMNN is a back propagating network that minimizes the error between the input and its reconstruction using gradient descent [16]. Learning parameter, weights and bias terms are initialized to very small random values to effectively learn different patterns. Training the input patterns to the specified accuracy (above 95%) is achieved by fixing a considerable amount of threshold distance in between the input and its mirror.

## 2. 3 Output Rescaling

As the modified hyperbolic tangent function itself results an output value in the range of -0.9 to +0.9 which is compatible with the truncated input, there is no need to rescale the desired output again at the time of training the patterns.

## 2.4 Pattern Classification using unsupervised learning

Automatic pattern classification of the training set as well as test set is done by simple partitional clustering technique which is widely used in engineering applications as described in [17]. We implemented this using Forgy's algorithm [18].

Forgy's algorithm:
1. Select initial seed points from input data set.
2. For all the input datasets repeat step 2.
   a. Calculate distance between each input data set and the each of the seed points representing a cluster.
   b. Place the input data into the group associated with the seed point which is closest to input data set (least of the distances in step 2 a)
3. Centroids of all the clusters are considered as new seed points.
4. Repeat step 2, 3 as long as the data sets leave one cluster to join another in step 2 b.

Input to the Forgy's algorithm is the dimensionally reduced central hidden layer's output. This is the feature set extracted from MMNN.

The number of initial selected seed points is equal to the number of distinguishable patterns in the training set. Initially, the seed points selected must be distinct in such a way that they perfectly cluster the input patterns and test patterns. The Euclidean distance between any two initial seed points belonging to two different categories is a representation of the inter-set distance between the two categories of images. The inter-set Euclidean distance is fixed by experimentation depending on the categories of images with which we train the mirroring neural network. We have used a threshold value 1.3 i.e., euclidean distances between any two initial seed points should be greater than 1.3.

For the initial seed points the following procedure is followed: First seed point is randomly selected from the input data set. Then, to select second seed point (from the input data set), the Euclidean distance between the first and the second must be greater than 1.3. Similarly to select $n^{th}$ seed point, the Euclidean distance between the present point and already selected n-1 seed points should be greater than 1.3. In our approach there are 3 categories of input data. So, we go for 3 initial seed points to separate the input into 3 distinguishable groups. After selecting the 3 initial seed points using the aforementioned procedure and initializing them as cluster centroids, find the cluster centroid nearest to each sample in the training set by comparing the Euclidean distances (distance between the input dataset and the cluster centroid). Mark the sample as of its nearest cluster. This is done for each of the samples in the training set (and/or test set). After completing the process of grouping the samples, compute the new centroid of the resulting clusters. Based on the new centroids of the resultant clusters, group the samples again by marking it as of its nearest cluster centroid. Repeat this procedure till there is no change in the cluster groups. This is done in unsupervised mode because while training the network we are not giving any information regarding the category of the input pattern, and classification is simply based on the selection of distant initial seed points.

**2.5 Results**

The inputs for training the multilayer mirroring neural network are 8-Bit grayscale images of size 26X26.

**Experiment 1:**
**(Two different input patterns; Face and Furniture)**

- While training MMNN, 176 images (88 face images and 88 furniture images) were given as input. MMNN was trained till it successfully mirrored 95% of the input images. With these weights and bias terms, the algorithm automatically classified the 176 images into two groups. The first group contained 89 images (88 furniture images and 1 face image).The second group contained 87 face images. From 176 input images, 175 images were classified correctly (99.43% success rate).
- The algorithm was tested with 80 new test images (40 face images, 40 furniture images) using the aforementioned weights and bias terms. These 80 images were entirely new and did not occur in the training set. The algorithm automatically classified the 80 test images into two groups. The first group contained 42 images (40 furniture images and 2 face images). The second group contained 38 face images. From a test sample space of 80 images 78 images were classified correctly (97.5% success rate).

**Experiment 2:**
**(Three different input patterns; Face, Flower and Furniture)**

- While training MMNN, 264 images (88 face images, 88 furniture images, 88 flower images) were given as input. MMNN was trained till it successfully mirrored 95% of the input images. With these weights and bias terms, the algorithm automatically classified the 264 input images into three groups. Group 1 contained 88 furniture images (i.e. for furniture samples 100% classification was achieved). Group 2 contained 84 flower images and 1 face image. Group 3 contained 87 face images and 4 flower images. From 264 input images, 259 images were classified correctly (98.1% success rate).
- The algorithm was tested with 90 new test images (30 face images, 30 furniture images and 30 flower images) using the aforementioned weights and bias terms. These 90 images were entirely new images not occurring in the training samples. The algorithm automatically classified the input images into three groups. The first group consisted of 30 furniture images (100% classification was achieved), second group consisted of 28 flower images. The third group consisted of 30 face and 2 flower images. From a test sample space of 90 images, 88 images were classified correctly (97.77% success rate).

We have also tested the mirroring neural network for generalized acceptance, learning competence and output reconstruction (please refer to the fig 2 for reconstruction).

The input and output images of the multilayer neural network are given in fig. 2.

**Inputs to the network**

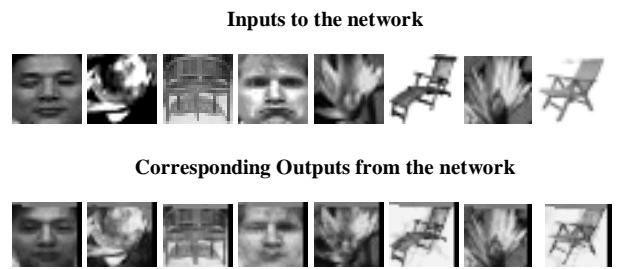

**Corresponding Outputs from the network**

**Fig. 2 Input and their corresponding output images**

## 3. Conclusion & Future Work

The algorithm proposed in this paper is a method of unsupervised learning and classification using mirroring neural networks and forgy's clustering technique.

To improve the performance of Forgy's clustering technique as applied to our application, we set a threshold distance between randomly selected initial seed points. This threshold made the randomly selected seed points sufficiently far apart as to make Forgy's technique cluster the input patterns perfectly.

The results of the algorithm over three different input patterns were encouraging. This can be extended to an architecture wherein the network will not only classify an input image pattern but it can also learn and classify any new pattern which is not one of the trained patterns.